\documentclass[10pt,twocolumn,letterpaper]{article}

\usepackage{cvpr}              % To produce the CAMERA-READY version
\usepackage{graphicx}
\graphicspath{{FIGURES/}}
\usepackage{amsmath}
\usepackage{amssymb}
\usepackage{booktabs}
\usepackage{diagbox}
\usepackage{helvet}
\usepackage{makecell}

\usepackage{mathrsfs}

\usepackage[pagebackref,breaklinks,colorlinks]{hyperref}
\usepackage[accsupp]{axessibility}

\usepackage[capitalize]{cleveref}
\crefname{section}{Sec.}{Secs.}
\Crefname{section}{Section}{Sections}
\Crefname{table}{Table}{Tables}
\crefname{table}{Tab.}{Tabs.}

\def\cvprPaperID{4109} % *** Enter the CVPR Paper ID here
\def\confName{CVPR}
\def\confYear{2022}

\begin{document}

\title{Infrared Invisible Clothing: \\Hiding from Infrared Detectors at Multiple Angles in Real World}

\author{Xiaopei Zhu$^{1, 2}$ \ \ Zhanhao Hu$^{2}$ \ \ Siyuan Huang$^{2}$  \ \  Jianmin Li$^{2}$ \ \ Xiaolin Hu$^{2,3,4}$\thanks{Corresponding author.}\\

$^{1}$School of Integrated Circuits, Tsinghua University, Beijing, China \\
$^{2}$Department of Computer Science and Technology, Institute for Artificial Intelligence, \\ State Key Laboratory of Intelligent Technology and Systems, BNRist, Tsinghua University, Beijing, China \\
$^{3}$IDG/McGovern Institute for Brain Research, Tsinghua University, Beijing, China \\
$^{4}$Chinese Institute for Brain Research (CIBR), Beijing, China \\
\tt\small \{zxp18, huzhanha17\}@mails.tsinghua.edu.cn \\ \tt\small \{siyuanhuang, lijianmin, xlhu\}@mail.tsinghua.edu.cn}

\maketitle
%%%%%%%%% ABSTRACT
\begin{abstract}
  Thermal infrared imaging is widely used in body temperature measurement, 
  security monitoring, and so on, but its safety research attracted attention only in recent years. We proposed the 
  infrared adversarial clothing, which could fool infrared pedestrian detectors at different angles. 
  We simulated the process from cloth to clothing in the digital world and then designed the adversarial 
  ``QR code" pattern. The core of our method is to design a basic pattern that can be expanded periodically, 
  and make the pattern after random cropping and deformation still have an adversarial effect, then we 
  can process the flat cloth with an adversarial pattern into any 3D clothes. 
  % We also proposed a new loss function: ${{L}_{black}}$, which aimed to lower the proportion of black pixels in 
  % the ``QR code" pattern and therefore minimize the heat-insulating material used in the physical clothing. 
  The results showed that the optimized ``QR code" pattern lowered the Average Precision (AP) of YOLOv3 by 87.7\%, while the random ``QR code" pattern 
  and blank pattern lowered the AP of YOLOv3 by 57.9\% and 30.1\%, respectively, in the digital world. We 
  then manufactured an adversarial shirt with a new material: aerogel. Physical-world experiments 
  showed that the adversarial ``QR code" pattern clothing lowered the AP of YOLOv3 by 64.6\%, while the random ``QR code" 
  pattern clothing and fully heat-insulated clothing lowered the AP of YOLOv3 by 28.3\% and 22.8\%, respectively. 
  We used the model ensemble technique to improve the attack transferability to unseen models. 
\end{abstract}

%%%%%%%%% BODY TEXT
\section{Introduction}
\label{sec:intro}

Thermal infrared (``infrared" for short throughout the paper) imaging is widely used in many areas such as human temperature measurement, safety monitoring, 
and autopilot. Infrared imaging has its unique advantages \cite{peikung_aaai}. First of all, infrared imaging can image in the dark, which means that the infrared equipment can work 24 hours a day. Secondly, unlike radar imaging, infrared imaging does 
not need to transmit signals actively, which saves more energy. Third, infrared 
imaging can measure temperature information, which is unavailable in visible light or radar filed. 
During the COVID-19 pandemic, infrared imaging is widely used for fast body temperature monitoring.

\begin{figure}[tb]
\centering
\includegraphics[width=0.8\columnwidth]{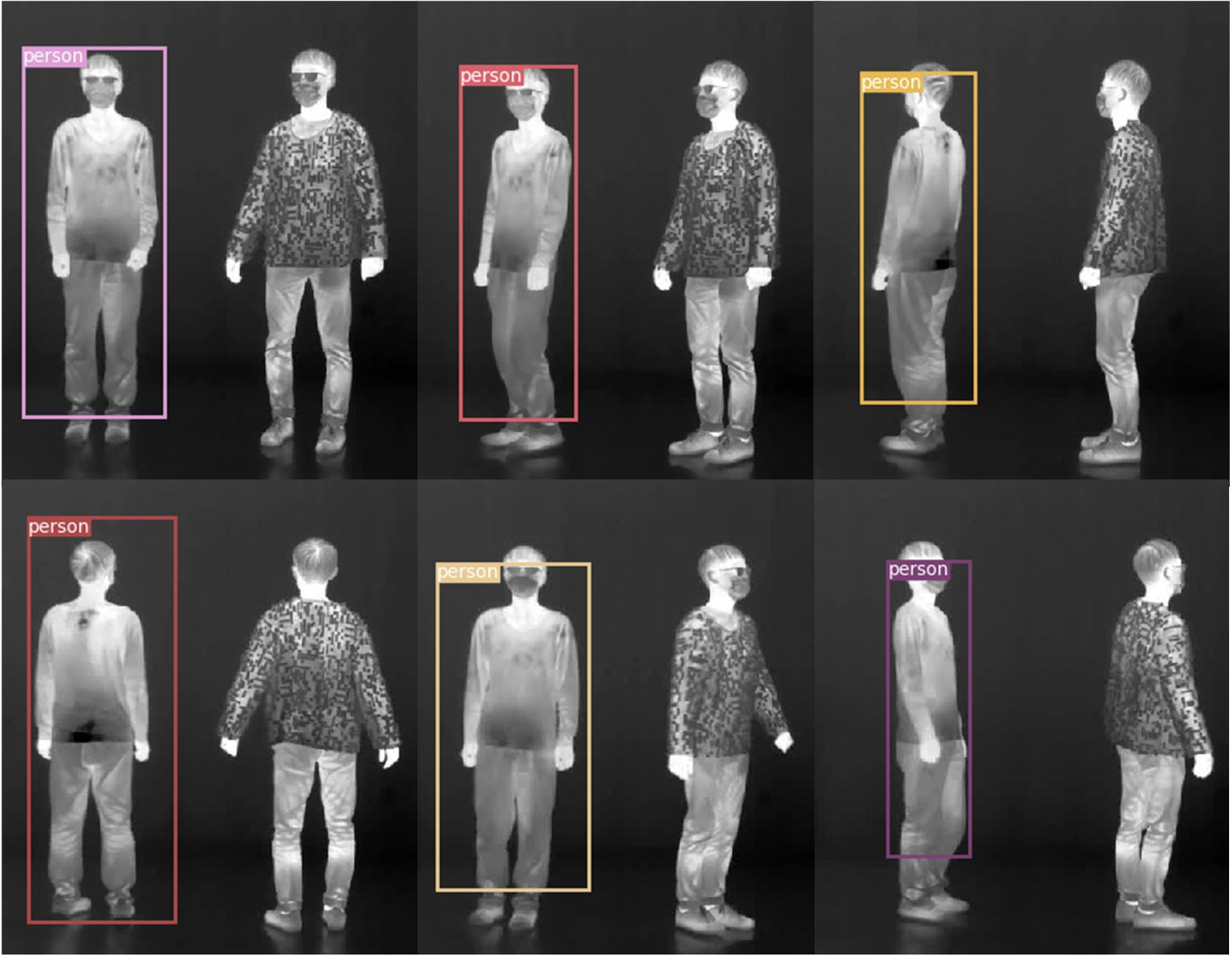} % lower the figure size so that it is slightly narrower than the column. Don't use precise values for figure width.This setup will avoid overfull boxes. 
\caption{Demonstration of physical infrared attack. A person wearing the adversarial clothes hid from infrared detectors at multiple angles. Whereas, the other person wearing ordinary clothes was detected (indicated by bounding boxes). }
\label{one_example}
\end{figure}

Infrared object detection combines deep learning with infrared imaging technique. The security of deep learning has 
attracted more and more attention in recent years. Szegedy et al. \cite{journals/corr/SzegedyZSBEGF13} found that 
neural networks can output error results with high confidence by adding specially crafted perturbations to the input 
data. The perturbed data is called adversarial example. Adversarial examples threaten not only the digital 
world \cite{journals/corr/SzegedyZSBEGF13,conf/sp/Carlini017,DBLP:journals/corr/GoodfellowSS14,conf/iclr/KurakinGB17,conf/ijcai/XiaoLZHLS18,conf/aaai/LiuLFMZXT19} 
but also the physical world \cite{conf/cvpr/EykholtEF0RXPKS18,conf/icml/AthalyeEIK18,conf/cvpr/ThysRG19,journals/corr/abs-1910-11099,conf/ccs/SharifBBR16}. 
Nowadays, most research on adversarial examples focuses on the visible light field; some are in the radar and infrared field. 
This paper focuses on the security of thermal infrared object detection systems.

Traditional infrared stealth generally uses two methods: heat insulation and active cooling, but they usually cannot 
completely hide thermal infrared signals. For example, as people need to breathe, the human body always emits thermal infrared 
signals to the outside. Adversarial example technology provides a different way of stealth, which can make deep learning-based 
detectors unable to detect people through carefully designed patterns \cite{conf/cvpr/ThysRG19,journals/corr/abs-1910-11099,conf/cvpr/HuangGZXYZL20}. 
Visible light patterns can be easily displayed in the physical world through printing or LED displays, but infrared patterns are difficult to be ``printed".

Zhu et al. \cite{peikung_aaai} proposed a physical method using small bulbs to attack infrared pedestrian detectors. 
To the best of our knowledge, that was the first work to realize physical attacks on the thermal infrared pedestrian detectors.  
But that method has an obvious shortcoming. The small bulb board can only attack at a specific angle (usually the front) 
of the human body. In this work, our goal is to solve this problem by designing a new physical attack method.  
Specifically, we want to design a piece of clothing to achieve a ``wearable" attack. There are two requirements for the adversarial clothing. First, the designed clothes should 
have a specific texture and deceive infrared pedestrian detectors from different angles. Second, the adversarial pattern on the clothing should still remain effective after non-rigid deformation.

Towards this goal, we designed infrared adversarial clothes based on a new material: \textit{aerogel}. 
The core of our method is to design a basic pattern that can be expanded periodically, and make the pattern after random cropping 
and deformation still have an adversarial effect, so that we can process the flat cloth with adversarial pattern into any 3D clothes.
Figure \ref{one_example} shows an example of physical infrared clothing attack and control experiments.
The contributions of this paper are as follows:

% \end{itemize}
\begin{itemize}
  \item First, we simulated the process from cloth to clothing in the digital world and then designed the adversarial ``QR code" pattern. 

  \item Second, we manufactured infrared adversarial clothing based on a new material aerogel, which hid from infrared detectors at multiple angles in the physical world.

\end{itemize}
%-------------------------------------------------------------------------
\section{Related Works}
\subsection{Digital Adversarial Attacks}
Since Szegedy et al. \cite{journals/corr/SzegedyZSBEGF13} discovered the vulnerability of deep neural networks, many digital attack methods 
have been proposed. Classic digital attack methods include gradient-based methods (e.g., FGSM \cite{journals/corr/GoodfellowSS14}, BIM \cite{conf/iclr/KurakinGB17}, DeepFool \cite{moosavi2016deepfool},
PGD \cite{conf/iclr/MadryMSTV18}), optimization-based methods (e.g., L-BFGS \cite{journals/corr/SzegedyZSBEGF13}, C\&W \cite{conf/sp/Carlini017}, ZOO \cite{DBLP:conf/ccs/ChenZSYH17}), and GAN-based methods (e.g., AdvGAN \cite{conf/ijcai/XiaoLZHLS18},
PS-GAN \cite{conf/aaai/LiuLFMZXT19}, AdvFaces\cite{conf/icb/DebZJ20}). The digital attack methods assume that the attacker can modify the model's input, 
which is difficult to achieve in the real-world setting. Some recent works \cite{conf/iclr/ZhangFDG19,journals/corr/abs-2006-14655} used 3D modeling to simulate 
real-world attacks, but there was still a gap between 3D modeling and real scenes.

\begin{figure*}[htbp]
\centering
\includegraphics[scale=0.66]{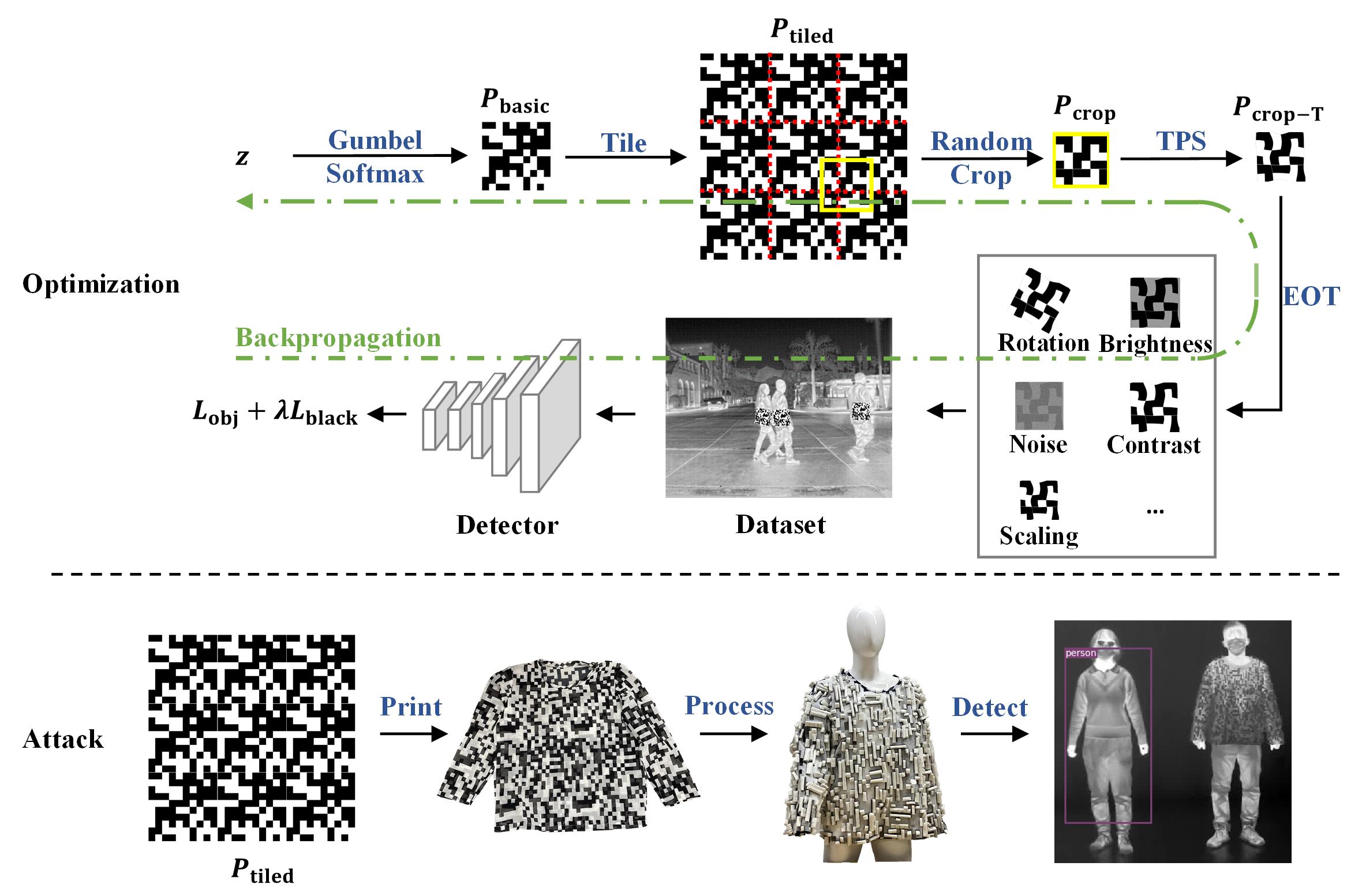} % lower the figure size so that it is slightly narrower than the column. Don't use precise values for figure width.This setup will avoid overfull boxes. 
\caption{Pipeline of proposed method. Top: attack in the digital world by optimizing a binary pattern. Bottom: attack in the physical world. }
\label{main_process}
\end{figure*}

\subsection{Physical Adversarial Attacks}

Most physical attacks focused on the visible light and the rader field. These attacks can be roughly divided into classification attacks and detection attacks. 
For classification attacks, Athalye et al. \cite{conf/icml/AthalyeEIK18} successfully deceived the classification model with a 3D printed tortoise. 
Eykholt et al. \cite{conf/cvpr/EykholtEF0RXPKS18} proposed Robust Physical Perturbations (RP2). Duan et al. \cite{conf/cvpr/DuanMQCYHY21} proposed Adversarial Laser Beam, which could quickly attack the classification 
model in the physical world in a non-contact manner. 

For detection attacks, Thys et al. \cite{conf/cvpr/ThysRG19} printed the adversarial pattern on 
a piece of paper and successfully made the YOLOv2 \cite{conf/cvpr/RedmonF17} unable to detect people. Xu et al. \cite{journals/corr/abs-1910-11099} designed 
a T-shirt with an adversarial pattern printed on the front. 
Huang et al. \cite{conf/cvpr/HuangGZXYZL20} proposed Universal Physical Camouflage (UPC) to fool Faster-RCNN \cite{journals/pami/RenHG017} in the physical world.  Hu et al. \cite{hu2021naturalistic} used the generative adversarial network (GAN) to generate natural 
looking adversarial patches while maintaining high attack performance. Tu et al. \cite{conf/cvpr/TuRMLYDCU20} proposed a method to generate 3D adversarial 
mesh to fool LiDAR detectors. A recent study \cite{conf/sp/CaoWXYFYCLL21} shows that the designed 
adversarial 3D-printed object could be invisible for both camera and LiDAR.

To the best of our knowledge, only one work focused on the safety of infrared object detection. Zhu et al. \cite{peikung_aaai}
designed a board decorated with small bulbs to attack infrared pedestrian detectors. The person holding 
the adversarial board could be invisible to the infrared detection model.

\subsection{Infrared Stealth Materials}
\label{sec:material}
Infrared stealth materials can be roughly divided into low emissivity materials and temperature control materials. Aluminum is a commonly used material with low emissivity,
but it is easily oxidized.
Fan et al. \cite{fan2020microwave} synthesized a new Al-reduced graphene oxide composite material, which had improved anti-oxidability and 
had excellent infrared stealth capabilities. 
% Rydzek et al. \cite{rydzek2012low} prepared aluminum-doped zinc oxide (AZO) films on glass substrates, and the infrared emissivity was 0.45. 
Temperature-controlled infrared stealth material realizes infrared stealth by reducing the surface temperature. 
Shang et al. \cite{shang2019microstructure} studied the microstructure and thermal insulation property of silica composite aerogel, 
which showed good thermal insulation stability at room temperature. 
Wang et al. \cite{wang2021polysiloxane} proposed a polysiloxane bonded silica aerogel with enhanced thermal insulation capability.

\section{Methods}
\subsection{Simulating the Cloth-to-Clothing Process in the Digital World}
\label{sec:simulation}
Our goal is to make a piece of clothing with adversarial texture in infrared imaging. It is required that this piece 
of clothing will have a certain adversarial effect from any angle. First of all, let's review the real-world 
process of making clothes. In the real world, we first look for a piece of cloth. We usually design a basic pattern and periodically expand 
the basic pattern on the plane. After we get the cloth printed with the expanded pattern, we crop and tailor it to clothing.

We simulate the process from the cloth to clothing in the digital world, 
as shown in Figure \ref{main_process}. Let ${{P}_{basic}}$ denote the basic pattern unit and ${{P}_{tiled}}$ denote the 
image after tiling (periodic expansion) of ${{P}_{basic}}$. ${{P}_{tiled}}$ can be any size.
% , which means that we generate a universal texture that is not limited to a fixed space. 
We define the tiling function as ${\rm TILE}$. 
The process above can be expressed as
\begin{equation}
  {{P}_{\text{titl}ed}}= {\rm TILE}\left( {{P}_{basic}} \right).
\end{equation}

When we take photos of people wearing clothes, we always photograph a part of the clothing, which can be 
regarded as cropping from the entire original cloth from which the clothing is made. We define a function ${\rm RC}$ for random cropping. The randomness here 
has two meanings: the randomness of the crop position and randomness of the crop size. 
Let ${{P}_{crop}}$ denote the patch randomly cropped from the 
pattern ${{P}_{tiled}}$, namely
\begin{equation}
  {{P}_{crop}}={\rm RC}\left( {{P}_{tiled}} \right).
\end{equation}

Due to the irregular deformation of cloth in the real world, we use the Thin Plate Spline (TPS) \cite{warps1989thin} interpolation 
method to approximate this process. TPS is an algorithm to simulate planar non-rigid deformation, especially 
suitable for cloth. 
% Figure \ref{TPS} shows an example of TPS transformation. 
Its basic idea is to give $K$ matching points in two images and make the points of one 
image with specific deformation correspond exactly to the points of the other image. Let ${{P}_{crop-T}}$ denote 
the patch after TPS transformation, namely
\begin{equation}
  {{P}_{crop-T}}={\rm TPS}({{P}_{crop}}).
\end{equation}

To simulate the disturbance in the real-world environment, like lighting changes, we used the Expectation over 
Transformation (EOT) \cite{conf/icml/AthalyeEIK18} method. 
\begin{equation}
  {{P}_{crop-TE}}={\rm EOT}\left( {{P}_{crop-T}} \right).
\end{equation}
EOT randomly changes the position, brightness, contrast, rotation angle, 
scale of the patch ${{P}_{crop-T}}$, and adds random noise to simulate changes in the real world as realistically as possible.

\subsection{Design of Binary Pattern}
\label{sec:gumbel}
We then consider the design of the cloth pattern. The mechanism of infrared imaging is quite different from that 
of visible light imaging. An infrared image is a grayscale image. The pixel value reflects the temperature 
of the object's surface. The higher the pixel value, the higher the temperature. We design the cloth 
pattern as shown in Figure \ref{main_process}, which looks like a Quick Response (QR) code. The white pixel reflects the normal body surface 
temperature, and the black pixel reflects the surface temperature after using thermal insulation materials. 
Therefore, we transform cloth pattern design into a binary optimization problem. For each pixel in the ``QR code", 
we only need to consider whether to put heat insulation material here or not. This design is beneficial to 
subsequent physical implementation.

However, since each pixel in the ``QR code" is a binary value, the ``QR code" pattern cannot be directly optimized 
by the gradient descent method. To solve this problem, we use the Gumbel-softmax technique \cite{conf/iclr/JangGP17}. The details 
are as follows. For each pixel in the image, ${{\pi }_{0}}$ denotes the probability of being black, 
and the probability of being white is ${{\pi }_{1}}$. 
% We use ${{\pi }_{i}}$ ($i=0,1$) to represent it uniformly. We already know that
Clearly, ${{\pi }_{0}}+{{\pi }_{1}}=1$. First, we introduce Gumbel noise. The purpose of adding 
Gumbel noise is to add randomness to the sampling operation. ${{g}_{i}}$ denotes the Gumbel noise of this pixel.
\begin{equation}
  {{g}_{_{i}}}=-\log \left( -\log \left( {{u}_{i}} \right) \right),{{u}_{i}}\sim Uniform\left( 0,1 \right).
\end{equation}
We next calculate the vector ${{y}_{i}}$ ($i=0,1$) used for sampling. $[{{y}_{0}},{{y}_{1}}]^\top$ is an approximate representation of the one-hot vector.
$\tau$ is a hyperparameter.  $[{{y}_{0}},{{y}_{1}}]^\top$ is closer to the one-hot vector when  $\tau$ is smaller. We choose $\tau=0.1$ in our experiment.
\begin{equation}
\begin{split}
   {{y}_{i}} & ={\rm Softmax} \left[ \left( {{g}_{i}}+\log {{\pi }_{i}} \right)/\tau  \right] \\ 
  & =\frac{\exp \left( \left( {{g}_{i}}+\log {{\pi }_{i}} \right)/\tau  \right)}{\sum\limits_{i=0}^{1}{\exp \left( \left( {{g}_{j}}+\log {{\pi }_{j}} \right)/\tau  \right)}}. \\ 
\end{split}
\end{equation}

\begin{figure}[tb]
\centering
\includegraphics[width=0.7\columnwidth]{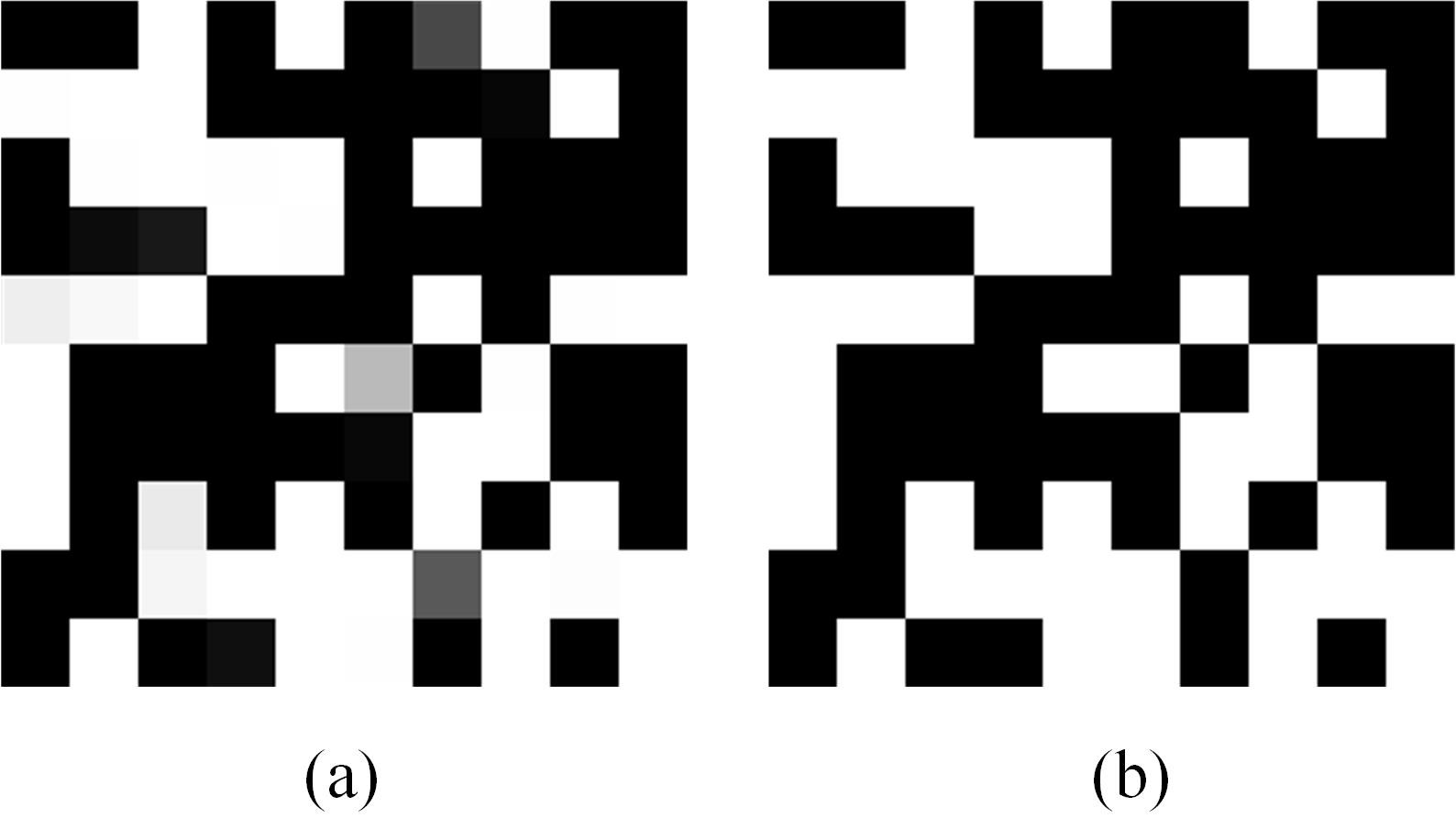} % lower the figure size so that it is slightly narrower than the column. Don't use precise values for figure width.This setup will avoid overfull boxes. 
\caption{The comparison between (a) the patch with approximate pixel values and (b) the patch with the real pixel values.}
\label{gumbel_convert}
\end{figure}
Next, we assume that the pixel value of the grayscale image is in the interval $\left[ 0,1 \right]$, and we can 
find the approximate value $\tilde{p}$ at that pixel, $\tilde{p}={{y}_{0}}\times 0+{{y}_{1}}\times 1={{y}_{1}}$. Since $\tilde{p}$ is 
differentiable relative to ${{\pi }_{i}}$, we can optimize $\tilde{p}$ using the gradient descent method. 
The corresponding relationship between the real value $p$ and the approximate value $\tilde{p}$ at this pixel is:
\begin{equation}
  p=\left\{ \begin{matrix}
    1,\tilde{p}\ge 0.5  \\
    0,\tilde{p}<0.5.  \\
 \end{matrix} \right. 
\end{equation}

In the optimization process, we use $\tilde{p}$ to approximate $p$ due to the need for gradient information; in the attack 
process, we use $p$ directly. Figure \ref{gumbel_convert} is a comparison between the patch with approximate pixel values and the 
patch with the real pixel values. It shows that the Gumbel-softmax technique can effectively help the approximation of binary images.

\subsection{Optimizing the Binary Pattern}
\label{sec:loss}
Figure \ref{main_process} shows the optimization process. ${{P}_{basic}}$ is an $N\times N$ patch. The variable $z$ in the 
hidden space is the set of probability values ${{\pi }_{i}} \left( i=0,1 \right)$ of each pixel in the patch
${{P}_{basic}}$, and the size of $z$ is $2\times N\times N$. ${{P}_{basic}}$ is tiled 
to ${{P}_{tiled}}$. ${{P}_{crop}}$ is randomly cropped from ${{P}_{tiled}}$. ${{P}_{crop-T}}$ is the patch 
after TPS transformation. ${{P}_{crop-TE}}$ is formed by ${{P}_{crop-T}}$ through EOT method. ${{P}_{crop-TE}}$ 
is pasted on the pedestrians in the data set, and then we input the patched images into the object detector. 
We update the variable $z$ according to the loss function and further update the patch ${{P}_{basic}}$.

Our loss function has two parts, ${{L}_{obj}}$ and ${{L}_{black}}$:
\begin{equation} 
  L={{L}_{obj}}+\lambda {{L}_{black}}. \label{v3_loss}
\end{equation}
The parameter $\lambda$ $(> 0)$ is determined empirically. We explain ${{L}_{obj}}$ and ${{L}_{black}}$ in what follows. 

${{L}_{obj}}$ denotes the object score of the object detector. Let $x$ denote the original image 
in the dataset and  $\tilde{x}$ denote the patched image. Let $f$ denote a model, $\theta $ denote its 
parameters. $f\left( x,\theta  \right)$ denotes the model's outputs given the input $x$. Most object detectors 
have three outputs: position of the bounding box ${{f}_{pos}}\left( x,\theta  \right)$, object 
probability ${{f}_{obj}}\left( x,\theta  \right)$ , and class 
probability ${{f}_{cls}}\left( x,\theta  \right)$. Our goal is to make object detectors unable to detect 
pedestrians, so we want to lower the ${{f}_{obj}}\left( x,\theta  \right)$ as much as possible. 
To fool the object detectors in the real world, we consider various transformations of the patches during 
attack, including translation, rotation, scale, noise, contrast, and brightness. Furthermore, we try to 
achieve a universal attack on different pedestrians due to the intraclass variety of pedestrians. 
Let $\mathbb{T}$ denote the set of transformations, ${{\tilde{x}}_{t}}$ denotes the patched image considering patch 
transformations. The data set has $M$ images. Considering all above factors, ${{L}_{obj}}$ can be described as
\begin{equation}
  {{L}_{obj}}=\frac{1}{M}\sum\nolimits_{i=1}^{M}{{{\rm E}_{t\in \mathbb{T}}}}f_{obj}^{(i)}\left( {{{\tilde{x}}}_{t}},\theta  \right).
\end{equation}

${{L}_{black}}$ is the average probability of black pixels appearing in patch ${{P}_{basic}}$. 
% Why do we propose such a loss function? Because 
The reason to propose this loss function is as follows. In physical implementation, the black pixels correspond to the 
heat-insulating material. The fewer black pixels are, the less heat-insulating material we use. This not 
only saves material, but also improves the air permeability and comfort of the clothes. From Section \ref{sec:gumbel}, 
we know that the probability of a single-pixel being black is ${{\pi }_{0}}$. For an $N\times N$ patch ${{P}_{basic}}$ , 
the average probability of black pixels is:
\begin{equation}
  {{L}_{black}}=\frac{\sum\limits_{i=0}^{N-1}{\sum\limits_{j=0}^{N-1}{{{\pi }_{0}}\left( i,j \right)}}}{N\times N}.
\end{equation}

% Therefore, the total loss function is 
% \begin{equation} 
%   L={{L}_{obj}}+\lambda {{L}_{black}}. \label{v3_loss}
% \end{equation}
% The parameter $\lambda $ is determined empirically.

For the ensemble attack \cite{conf/iclr/LiuCLS17}, we aim to lower each detector's objectness score at the same time. We assume there 
are $F$ detectors, and the objectness score of $i$-th detector is $L_{obj}^{(i)}$. We take the sum of these losses. Thus, the total 
loss of the ensemble attack is
\begin{equation}
  {{L}_{ensemble}}=\sum\limits_{i=1}^{F}{L_{obj}^{(i)}}+ \lambda{{L}_{black}}. \label{ensem_loss}
\end{equation}

\subsection{Physical Implementation}
\label{sec:physical implementation}

Infrared stealth materials can be roughly divided into low emissivity materials and temperature control materials.
According to the Stefan-Boltzmann law \cite{de1995generalized}, the infrared radiation is more sensitive to temperature,
so we gave priority to temperature control materials. 
We tested two common fabrics (cotton and polyester), 
two thermal insulation tapes (Teflon and polyimide), and a new type of material (aerogel). See \textit{Supplementary Material} for their photos. 
We aimed to find a material that has the best thermal insulation performance. 

The process for making clothes is as follows. First, we printed the "QR code" pattern we designed on 
a $1.5m\times1.5m$ cloth. Next, We hired a tailor to make the cloth into a piece of clothing. 
Then we cropped the infrared stealth material into blocks and stuck them on the black area of the clothes.

\section{Experiments}
\subsection{Dataset}

We used the $FLIR\_ADAS\_v1\_3$ dataset \cite{FLIR_ADAS} released by FLIR company. $FLIR\_ADAS\_v1\_3$ provides an annotated thermal image set for 
training and validation of object detection. The original dataset contains four types of objects, namely people, dogs, cars and bicycles. 
Since we focused on people, we filtered the dataset and selected 9900 images that contained people. We named the subset $PEOPLE\_FLIR$. 
The training set contained 7873 images, and the test set contained 2027 images. 

\subsection{Target Detector}
We followed the work of Zhu et al. \cite{peikung_aaai} and chose the same target detector YOLOv3 \cite{journals/corr/abs-1804-02767} . Kristo et 
al. \cite{journals/access/KristoIP20} compared the performance of state-of-the-art infrared detectors such as Faster-RCNN \cite{ren2016faster}, 
Cascade-RCNN \cite{cai2018cascade}, SSD \cite{liu2016ssd}, and YOLOv3. They found that YOLOv3 was significantly faster than other 
detectors while achieving performance comparable with the best. We resized the input images to 
416 $\times$ 416 as required by YOLOv3. 
We used the pretrained weights officially provided by YOLO and then fine-tuned them on $PEOPLE\_FLIR$. The target model's AP 
was 97.27\% on the training set and 85.01\% on the test set. In our experiments, we first attacked YOLOv3 and then  
attacked other detectors under the black box setting.

\subsection{Simulation of Physical Attacks}
\subsubsection{Attack YOLOv3 in the Digital World}
\label{sec:digital attack}
As mentioned in Section \ref{sec:loss}, the size of variable $z$ is $2\times N\times N$. ${{P}_{basic}}$ is an $N\times N$ patch. 
In our experiment, we chose $N=20$ (See section \ref{sec:patch size} for the results of other 
values). We expanded the side length 
of ${{P}_{basic}}$ by 5 times, so the size of ${{P}_{tiled}}$ was 100$\times$100. ${{P}_{crop}}$ was 
randomly cropped from ${{P}_{tiled}}$, the crop size was randomly sampled 
from [10,30]. We took matching points $K = 16$ in TPS transformation. 
The set of EOT transformations included changes of patch position, brightness, contrast, rotation, angle, scale, and noise.

\begin{figure}[tb]
\centering
\includegraphics[width=0.7\columnwidth]{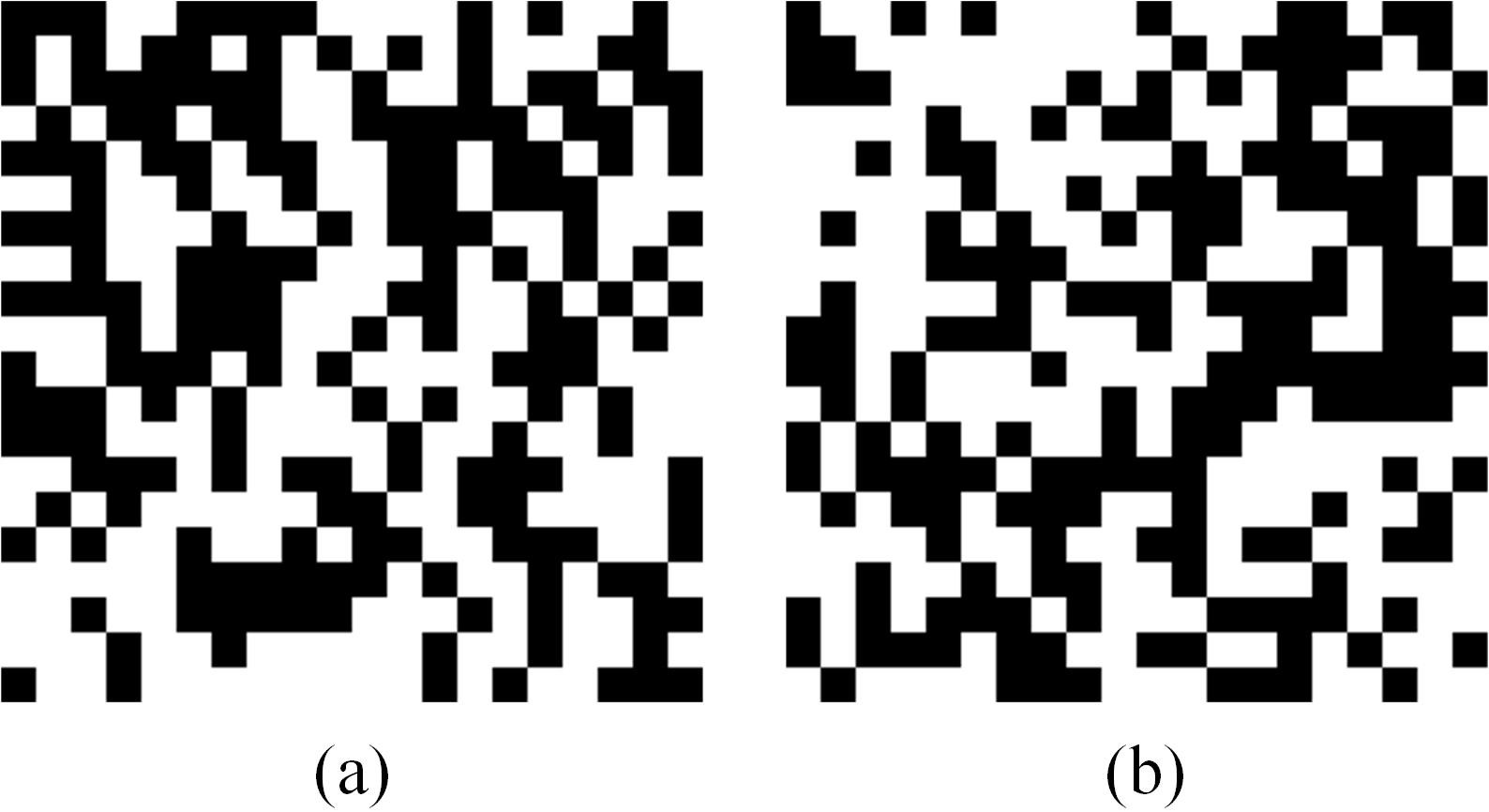} % lower the figure size so that it is slightly narrower than the column. Don't use precise values for figure width.This setup will avoid overfull boxes. 
\caption{Optimized ``QR code" texture based on (a) YOLOv3 (b) ensemble models.}
\label{digital_yolov3}
\end{figure} 

\begin{figure*}[htbp]
\centering
\includegraphics[scale=0.45]{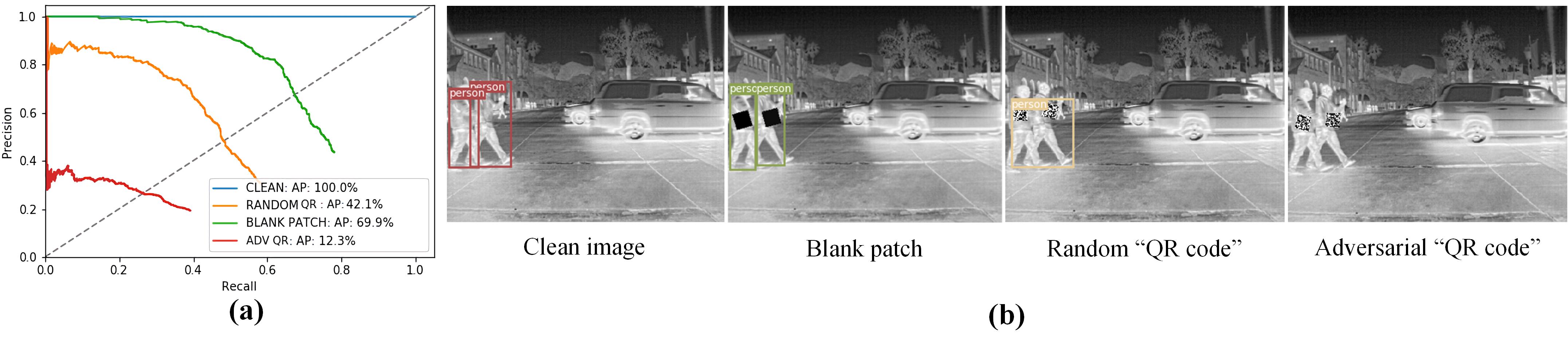} % lower the figure size so that it is slightly narrower than the column. Don't use precise values for figure width.This setup will avoid overfull boxes. 
\caption{Digital attacks. (a) Evaluation of digital attacks. (b) Examples of digital attacks. Bounding boxes indicate successful detecting of persons.}
\label{digital_examples}
\end{figure*}
Next, we used the training set of $PEOPLE\_FLIR$, and placed ${{P}_{crop-TE}}$ in a random 
position of the human body according to the bounding box. The proportion of the patch size to the 
height of the bounding box varied from 0.1 to 0.3 according to the crop size. Next, we inputted 
these patched images into YOLOv3. We used a stochastic gradient optimizer with momentum. The 
optimizer used the backpropagation algorithm to update the parameters of variable $z$ by minimizing 
Equation \ref{v3_loss}, and further updated the patch ${{P}_{basic}}$. The hyper-parameter $\lambda$ in loss 
function was 0.1 (the sensitivity of this parameter is analyzed in Section \ref{sec:lambda}). 
See \textit{Supplementary Material} for details about the hyperparameter setting such as batch size, learning rate, etc.
Figure \ref{digital_yolov3}(a) shows the optimization result.

Next, we applied the optimized pattern (Figure \ref{digital_yolov3}(a))  to the test set in the same way as the 
optimization process. To further compare the effect of the attack, we used a random ``QR code" pattern 
and a blank pattern for control experiments. In our experiment, the pixel value of the blank pattern 
was 0, which corresponded to the situation where the heat was completely insulated in the real world. We applied these patterns 
to the test set of $PEOPLE\_FLIR$, and then inputted the patched images to YOLOv3 to test their 
attack performance. We defined the model's output of clean images input as ground truth (GT). We 
used the Intersection over Union (IOU) method to calculate the detection accuracy. The 
precision-recall (P-R) curves are shown in Figure \ref{digital_examples}(a). The results showed that the optimized ``QR code" 
pattern made the average precision (AP, the area under the PR curve) of YOLOv3 drop by 87.7\%. 
In contrast, the random ``QR code" pattern and blank pattern made the AP of YOLOv3 drop 
by 57.9\% and 30.1\%, respectively. Although random ``QR code" pattern and blank pattern 
also lowered AP of the model, it is far less effective than the optimized pattern. Figure \ref{digital_examples}(b) shows some examples in the digital world.

\begin{table}[bp]\small %\footnotesize
\caption{Effect of Resolution of Basic Patch}\label{tab_patch_size}
\centering
% \setlength\tabcolsep{3.5pt}
% \caption{Study of number of Gaussian functions}\label{tab1}
\begin{tabular}{cccccc}
\toprule[1.1pt]  
Patch size & 10 & 20 & 30 & 40 & 50\\
\midrule  
AP decrease &  64.9\% & 87.7\% & 85.6\% & 83.6\% & 83.4\%\\
\bottomrule[1.1pt] 
\end{tabular}
\end{table}

\subsubsection{Effect of Resolution of the Basic Patch}
\label{sec:patch size}
In the previous experiment (Section \ref{sec:digital attack}), the resolution of ${{P}_{basic}}$ was 20$\times$20. 
% So if we choose a different resolution of basic patch, how will it affect the result? 
We then studied the resolution of 10$\times$10, 20$\times$20, 30$\times$30, 40$\times$40 
and 50$\times$50. The pattern optimization and testing process were described in Section \ref{sec:digital attack}. Table \ref{tab_patch_size} showed the AP decrease of YOLOv3 corresponding to 
different resolutions. The more AP decrease meant the better attack effect. The result showed 
that the attack performance decreased if the resolution was too large or too small. And 20$\times$20 
was our best result.

\subsubsection{Effect of the Parameter $\lambda$ in the Loss Function}
\label{sec:lambda}
The parameter $\lambda$ in Equation \ref{v3_loss} balances ${{L}_{obj}}$ and ${{L}_{black}}$. 
We studied lambda values of 0, 0.01, 
0.05, 0.1, 0.5, and 1. We adopted the same optimization and testing methods as described in Section \ref{sec:digital attack}. We 
evaluated the attack performance of patterns generated with different $\lambda$ values by AP decrease. 
We also calculated the proportion of black pixels in different patterns. The results are shown in 
Table \ref{tab_labda}. The results showed that there was a certain trade-off between improving the attack 
performance and reducing the proportion of black pixels in the basic patch. The higher the proportion 
of black blocks, the stronger the attack performance of patterns under certain conditions.

\begin{table*}[htbp]\small
\caption{Effect of Parameter $\lambda$ in the Loss Function}\label{tab_labda}
\centering
% \setlength\tabcolsep{3.5pt}
% \caption{Study of number of Gaussian functions}\label{tab1}
\begin{tabular}{ccccccc}
\toprule[1.1pt]  
$\lambda$ & 0 & 0.01 & 0.05 & 0.1 & 0.5 & 1\\
\midrule  
Black pixel ratio&  52.0\% & 49.8\% & 48.2\% & 47.3\% & 45.3\% & 44.3\%\\
AP decrease&  89.2\% & 88.4\% & 88.2\% & 87.7\% & 87.6\% & 87.2\%\\
\bottomrule[1.1pt] 
\end{tabular}

\end{table*}

\begin{figure*}[htbp]
\centering
\includegraphics[scale=0.18]{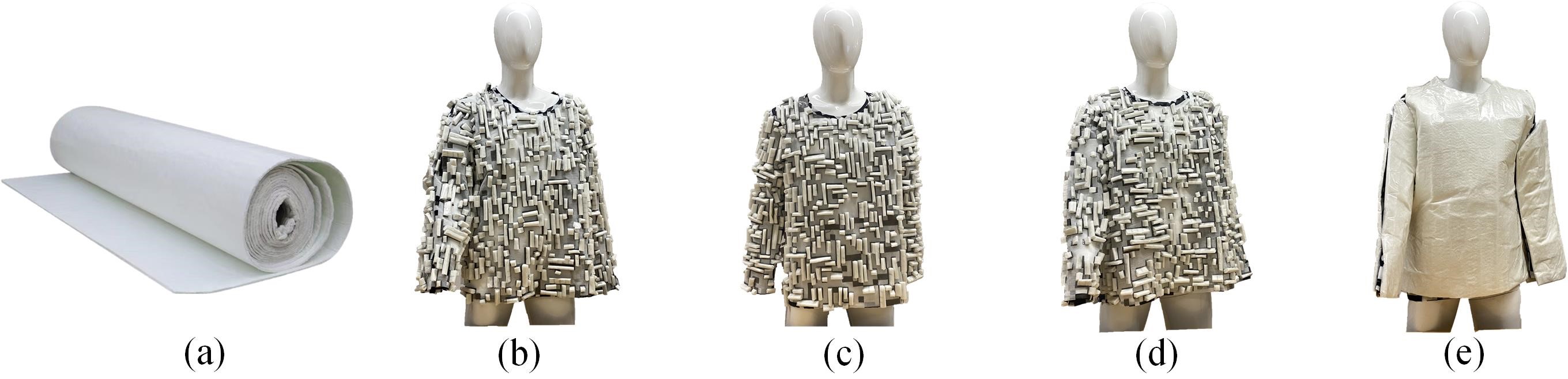}
\caption{Physical material and clothing. (a) Aerogel felt.  (b) Adversarial clothing based on YOLOv3.  (c) Random ``QR code'' clothing. (d) Adversarial clothing based on ensemble models. (e) Fully heat-insulated clothing.}
\label{phy_cloth}
\end{figure*}

\subsection{Attacks in the Physical World}
\subsubsection{Physical Test of Thermal Insulation Materials}
We tested the thermal insulation performance of the five materials stated in Section \ref{sec:physical implementation}. See \textit{Supplementary Material} for more details.
The results showed that the aerogel had good thermal insulation properties and remained stable over time.

\begin{figure*}[htbp]
\centering
\includegraphics[scale=0.56]{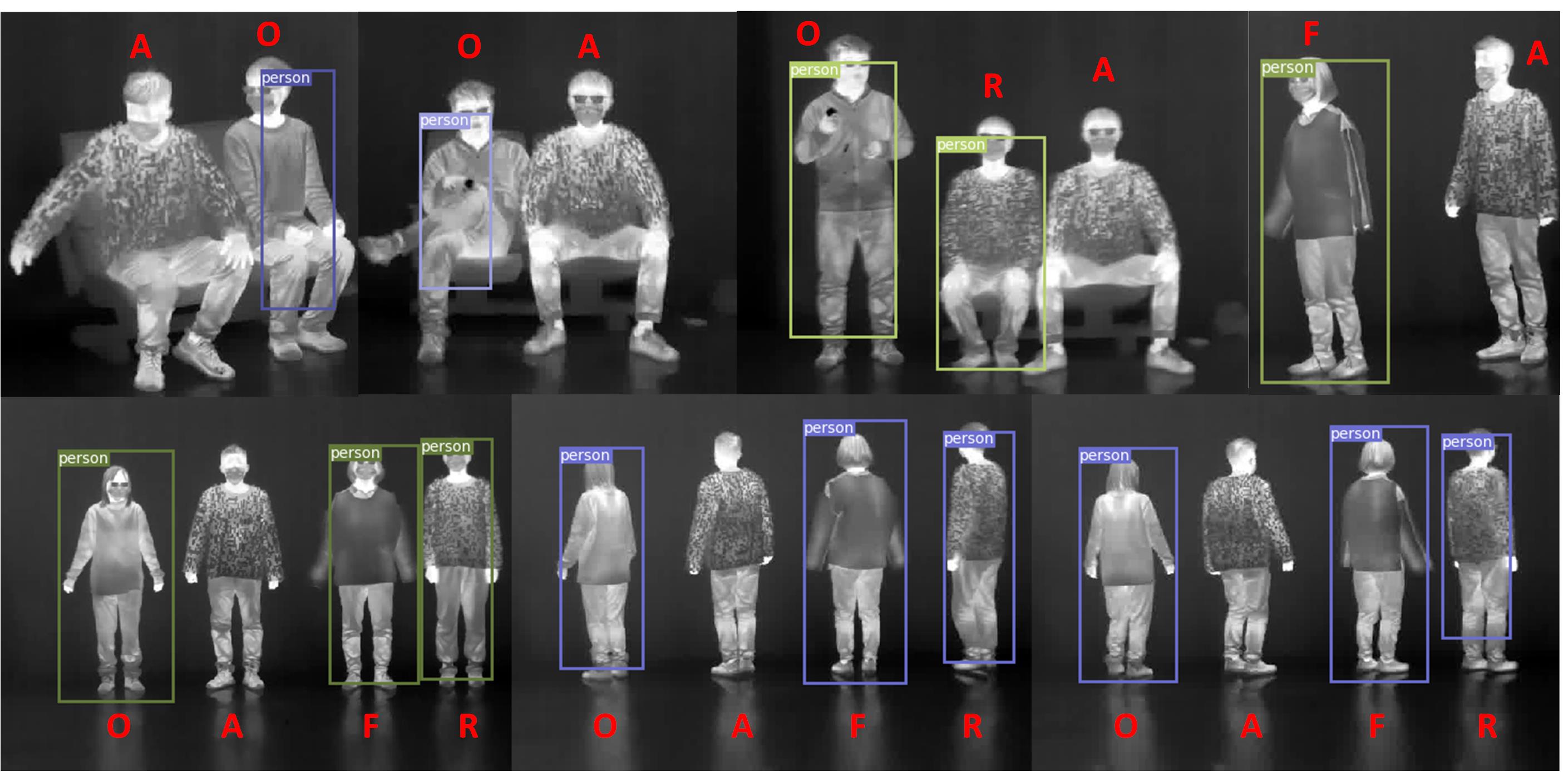}
\caption{Visulization results of physical attacks. Persons wearing different clothing could be in various poses. A: adversarial ``QR code'' clothing. R: random ``QR code'' clothing. F: fully heat-insulated clothing. O: ordinary clothing.}
\label{more_exapmples}
\end{figure*} 

\subsubsection{Attack YOLOv3 in the Physical World}
\label{sec:physical attack}
% Section \ref{sec:physical implementation} introduced the process from the optimized digital pattern to the physical clothing.
% The finished clothing is shown in Figure \ref{phy_cloth}. 
% For better comparison, we also made a piece of random ``QR code" pattern clothing. These clothes had the same size. Next, we tested the attack performance of these clothes 
% in real world.
We cropped the aerogel felt (Figure \ref{phy_cloth}(a)) into many blocks and stuck them on the clothes. The manufacturing process is in the \textit{Supplementary Video}. The manufacturing 
cost of our whole piece of clothing was within 50 USD, which meant the possibility of mass production. The finished clothing was shown in Figure \ref{phy_cloth}(b). For better comparison, we also made a piece of random ``QR code" pattern 
clothing (Figure \ref{phy_cloth}(c)) and a piece of fully heat-insulated clothing (Figure \ref{phy_cloth}(e)). These clothes had the same size. Next, 
we tested the attack performance of these clothes in the real world.

The infrared camera we used was FLIR T630sc (FPA 640 $\times$ 480, NETD$<$40mK). We invited 7 
volunteers to participate in our experiment. This experiment was approved by the Institutional Review Board (IRB). The volunteers wore adversarial ``QR code" clothing,  
random ``QR code" clothing, fully heat-insulated clothing or ordinary clothing. 
We photographed volunteers in multiple scenes indoors and outdoors, and the camera's distance 
from them varied within 1-15 meters. At the same time, they can change different postures according 
to their preferences, such as standing, sitting, and even reclining, etc. We photographed the 
volunteers in different scenes simultaneously and sent these infrared images to YOLOv3. The 
threshold of the detection output was 0.7. Figure \ref{more_exapmples} gives some specific 
examples. We can see that people wearing adversarial ``QR code" clothing were not detected by 
YOLOv3 even if they were in different postures, keeping different distances from the camera, 
and in different scenes. While at the same time, people wearing random "QR code" clothing,
fully heat-insulated clothing or ordinary clothing were detected. The results showed 
the effectiveness of our method in the physical world. See \textit{Supplementary Video} for the demo.

To quantitatively evaluate the effect of physical attacks, we recorded 120 videos in different scenes. 
60 videos were recorded indoors, and the others were recorded outdoors. 5 volunteers were the actors 
in the video. We fixed the camera's position. Then we selected three typical positions which were 
3 meters, 5 meters, and 7 meters from the camera to test attack performance. 
Then we invited volunteers to rotate in situ at a constant speed in these positions. For fair comparison, the same volunteer needed to wear 
adversarial ``QR code" clothing, random ``QR code" clothing, fully heat-insulated clothing and ordinary 
clothing, respectively, in the same position. We recorded videos with our infrared camera. We sampled 
the videos (from 5 volunteers) at 3 frames per second, and got 900 frames per condition, which made 3600 frames in total. 
We used manual annotation as the ground truth (GT) 
and then used the output of YOLOv3 to calculate AP by IOU method. The results showed that the 
adversarial ``QR code" clothing reduced the detector's AP by 64.6\%, while the random 
``QR code" clothing, fully heat-insulated clothing and ordinary clothing made the detector's AP 
drop by 28.3\%, 22.8\% and 4.0\%, respectively. 

\begin{figure}[tb]
\centering
\includegraphics[width=0.9\columnwidth]{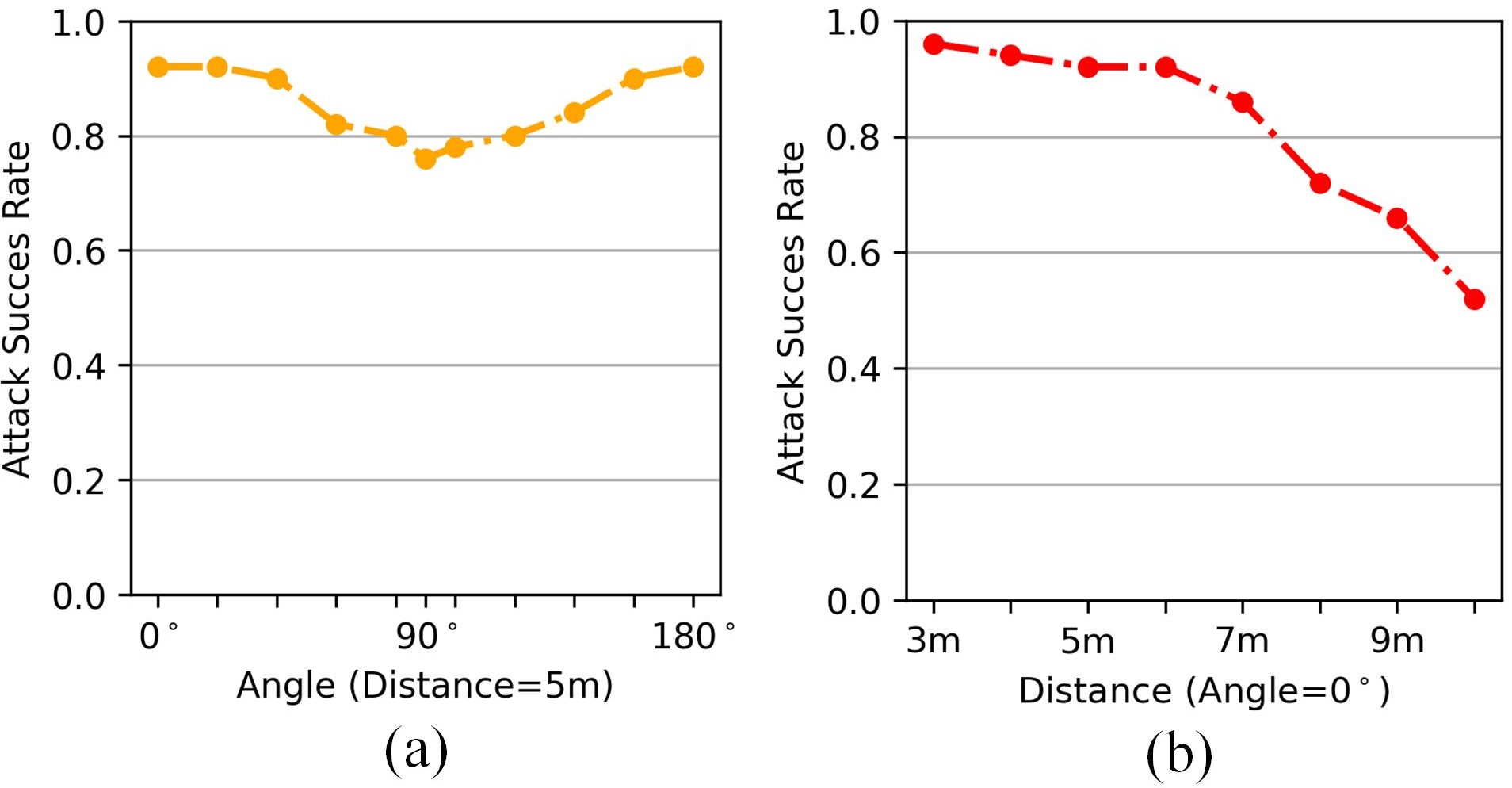} 
\caption{Analysis of attacks at different (a) angles and (b) distances}
\label{angle_distance}
\end{figure} 

% \subsubsection{Analysis of Attacks at Different Angles and Distances}
% \label{sec:angle}
We then analyzed the attack effect at different angles.
We chose a distance of 5m from the camera, and marked some key angles on the ground. We then took 11 
sample points from a counterclockwise rotation angle of $0^{\circ}$ to $180^{\circ}$ for statistics. We selected 100 frames 
at each angle, and used the attack success rate (ASR, the ratio of the number of frames that were not 
detected to the total number of frames) as the evaluation method. The threshold of the detector 
was 0.7. As shown in Figure \ref{angle_distance}(a), we plot the curve of the ASR with different angles.  
When the volunteers had their front to the camera ($0^{\circ}$) or back to the camera ($180^{\circ}$), the ASR was the highest. When they had their side to the camera 
($90^{\circ}$), the ASR was the lowest. One possible reason is that the area of the adversarial pattern on the 
side ($90^{\circ}$) was relatively small. 

We then studied the relationship between the ASR and distances. We kept the clothes always towards ($0^{\circ}$) the camera, and the 
distance from the camera varied from 3 to 10 meters. We took 8 sample points between 3 and 10 meters for statistics, and selected 100 
frames at each position. We plot the curve of the ASR with different distances (Figure \ref{angle_distance}(b)). 
When the distance was between 3 and 7 meters, the ASR was above 0.8. When the distance exceeded 7 meters, the ASR dropped faster, 
because the adversarial pattern was not easy to attack successfully in a smaller view.

\subsection{Ensemble Attack}
\label{sec:ensemble}
We found attack transferability of the single model was limited. The pattern optimized on 
YOLOv3 only lowered the AP of Deformable DETR, RetinaNet, and Libra-RCNN by 13.7\%, 25.3\%, and 32.7\% in 
the digital world, respectively. We then used the model ensemble technique \cite{conf/iclr/LiuCLS17} as described in Section \ref{sec:loss}.
Figure \ref{digital_yolov3}(b) shows a pattern obtained by ensembling YOLOv2, YOLOv3, Faster-RCNN, and Mask-RCNN \cite{he2017mask} during the optimization 
process. It caused the AP of Deformable DETR, RetinaNet, and Libra-RCNN to drop by 24.7\%, 58.2\%, and 66.9\% in the digital world. 
Then we manufactured a piece of clothing with the pattern obtained by model ensembling in the physical world (see Figure \ref{phy_cloth}(d)), which 
made the AP of Deformable DETR, RetinaNet, and Libra-RCNN drop by 16.2\%, 40.4\%, 51.9\%, respectively. Details can be found in \textit{Supplementary Material}.

\subsection{Adversarial Defense Methods}
We tested five typical methods to defend our attack method in the digital world. 
These methods included preprocessing defenses (spatial smoothing \cite{conf/ndss/Xu0Q18} and Total Variance Minimization \cite{conf/iclr/GuoRCM18}, 
adversarial training \cite{journals/corr/GoodfellowSS14}, and their combinations. 
The most effective way increased the AP from 12.3\% to 36.8\% only,
and our attack method still lowered the AP by 63.2\%. See \textit{Supplementary Material} for details.

\section{Conclusion and Discussion}

\noindent \textbf{Summary.   }  This paper presents a new method of design and manufacturing infrared adversarial clothing. We simulated 
the process from cloth to clothing in the digital world and then designed the adversarial "QR code" pattern. 
We manufactured infrared adversarial clothing based on a new material 
aerogel. Compared with the small bulbs board \cite{peikung_aaai}, our adversarial clothing hid from infrared detectors from multiple angles.

\noindent \textbf{Limitations.  } As mentioned in Section \ref{sec:physical attack}, the adversarial clothing had a significant 
decrease in the ASR when it was far away from the camera.  As mentioned in Section \ref{sec:ensemble}, it is seen that the adversarial patterns were difficult to attack the transformer-based model,
since our adversarial patterns were generated based on the CNN models.
 
\noindent \textbf{Potential Negative Impact.  }  Adversarial example techniques should be used carefully. If abused, adversarial attacks may 
threaten the security of AI systems. However, adversarial attack also promotes the research of defense methods.

\section{Acknowledgements}
This work was supported in part by the National Natural Science Foundation of 
China (Nos. U19B2034, 62061136001, 61836014) and the Tsinghua-Toyota Joint Research Fund.

%%%%%%%%% REFERENCES
{\small
\bibliographystyle{ieee_fullname}
\bibliography{egbib}
}

\end{document}